\documentclass[letterpaper, 10pt,conference,twocolumn]{ieeeconf}

\usepackage[ruled,linesnumbered]{algorithm2e}
\usepackage{graphicx}
\usepackage{float}

\IEEEoverridecommandlockouts  
\usepackage{times,multirow,caption,float}
\usepackage{wrapfig}
\usepackage{graphicx}
\usepackage{microtype}
\usepackage{svg}
\usepackage{epsfig,xspace,layout}
\usepackage{subfigure}
\usepackage{color}
\usepackage{amsfonts}
\usepackage{times}
\usepackage{amssymb}
\usepackage{amsmath}
\usepackage{rotating}

\usepackage{amsthm}
\usepackage{graphicx}
\usepackage{epsfig}
\usepackage{mathrsfs}
\usepackage{caption}
\usepackage{subfigure}
\usepackage{mathtools}
\usepackage{makeidx}
\usepackage{multirow} 
\usepackage{dblfloatfix}
\usepackage{threeparttable}
\usepackage{dsfont}
\usepackage[font={small}]{caption}
\usepackage{cleveref}
\usepackage{tikz}
\usetikzlibrary{shapes,arrows,positioning,calc}
\usepackage{siunitx}
\usepackage{url}
\usepackage{cite}

\theoremstyle{definition}

\newtheorem{problem}{Problem}
\theoremstyle{remark}
\newtheorem{remark}{Remark}

\DeclareMathAlphabet{\mathpzc}{OT1}{pzc}{m}{it}

\DeclareFontFamily{U}{jkpmia}{}
\DeclareFontShape{U}{jkpmia}{m}{it}{<->s*jkpmia}{}
\DeclareFontShape{U}{jkpmia}{bx}{it}{<->s*jkpbmia}{}
\DeclareMathAlphabet{\mathfrak}{U}{jkpmia}{m}{it}

\newcommand{\MSE}{\text{MSE}}

\newcommand{\0}{\mathbf{0}}
\newcommand{\R}{\mathbb{R}}

\newcommand{\V}{\mathcal{V}}

\newcommand{\W}{\mathcal{W}}

\newcommand{\F}{\mathcal{F}}

\renewcommand{\P}{\mathcal{P}}

\newcommand{\eq}[1]{\begin{equation}#1\end{equation}}
\newcommand{\ald}[1]{\begin{aligned}#1\end{aligned}}

\newcommand{\eqn}[1]{\begin{equation*}#1\end{equation*}}

\begin{document}
\title{Polytopic Planar Region Characterization of Rough Terrains for Legged Locomotion}
\author{Zhi Xu, Hongbo Zhu, Hua Chen, and Wei Zhang
\thanks{The authors are with Department of Mechanical and Energy Engineering, Southern University of Science and Technology, Shenzhen, China. Emails: {\tt \{11812426,zhuhb\}@mail.sustech.edu.cn, \{chenh6, wzhang3\}@sustech.edu.cn}}}
\maketitle

\begin{abstract}
This paper studies the problem of constructing polytopic representations of planar regions from depth camera readings. This problem is of great importance for terrain mapping in complicated environment and has great potentials in legged locomotion applications. To address the polytopic planar region characterization problem, we propose a two-stage solution scheme. At the first stage, the planar regions embedded within a sequence of depth images are extracted individually first and then merged to establish a terrain map containing only planar regions in a selected frame. To simplify the representations of the planar regions that are applicable to foothold planning for legged robots, we further approximate the extracted planar regions via low-dimensional polytopes at the second stage. With the polytopic representation, the proposed approach achieves a great balance between accuracy and simplicity. Experimental validations with RGB-D cameras are conducted to demonstrate the performance of the proposed scheme. The proposed scheme successfully characterizes the planar regions via polytopes with acceptable accuracy. More importantly, the run time of the overall perception scheme is less than 10ms (i.e., $> 100$Hz) throughout the tests, which strongly illustrates the advantages of our approach developed in this paper.
\end{abstract}

\section{Introduction}
\label{Introduction}
The ability of perceiving the environment and building structured maps that can be used in path and motion planning is among the most critical abilities of mobile robots. For legged robots, due to the discretely changing footholds, such an ability becomes even more vital. The most significant advantage of legged robots as compared with traditional wheeled ones is their adaptability to complex terrains. Obviously, being able to extract terrain information is a premise of realizing such advantages. Hence, investigations on terrain understanding and characterization are of great importance for fully exploiting legged robots.

The problem of establishing usable maps for path and trajectory planning for mobile robots has been extensively studied in both robotics and computer vision communities. Especially, the simultaneous localization and mapping (SLAM) has long been one of the most hottest topics in robotics~\cite{lee_indoor_2012,geneva_lips_2018,yang_monocular_2019,zuo_lic-fusion_2020,sun_plane-edge-slam_2021}. Three dimensional (3D) reconstruction, which is a classical problem in computer vision that has been attracting considerably increasing research attentions recently~\cite{yun_3d_2007,keller_real-time_2013,liu_3d_2016,pan_dense_2019}, is fundamentally of the same mathematical nature. The specific terrain mapping problem for legged robots differs from the above standard problems in various aspects. Among them, the requirement of providing guidance on how to selected future foothold is perhaps the most distinctive feature. How to accurately, reliably and efficiently establish a structured map that encodes this feature has not been given adequate research attention. 
\begin{figure}[t]
\centering
\includegraphics[height=5.5cm]{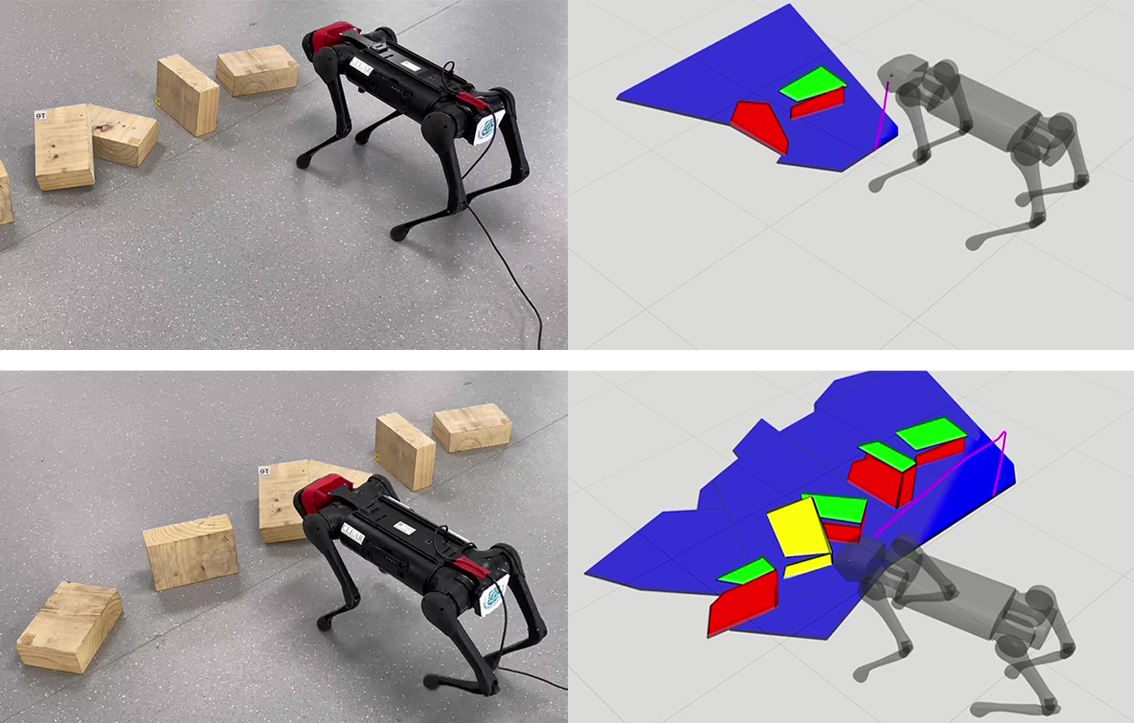}
\caption{{\small Ploytopic planar region characterization results.}}
\label{fig:fig_one}
\end{figure}

Trying to bridge the aforementioned gap, we focus on the planar region extraction and polytopic approximation problem in this paper. Inspired by the fact that regions of candidate footholds serve as constraints in foothold planning problem of legged robots, we aim to construct a terrain map consisting of only polytopes, which renders the constraints linear and therefore amenable. Given a sequence of depth images and the associated camera frames, our proposed approach first segments the candidate planes within each depth image. Then, planar regions extracted from different depth images that are actually corresponding to the same physical plane are merged to respect the integrity of the real planar regions. Once the pixel-wise planar region characterizations are constructed, we develop a polygonal approximation to balance between accuracy and tractability. Finally, the potentially non-convex polygons are convexified via polytopic partitions to generate the desired polytopic representation of the terrain.




\subsection{Related Works}


As briefly mentioned above, 3D reconstruction and Simultaneous Localization And Mapping (SLAM) both aim to generate representations of the environment, which lay the foundation for more specialized terrain mapping strategies. Investigations on 3D reconstruction mainly aim to establish the precise reconstruction of the 3D model of objects or scenes \cite{yun_3d_2007,keller_real-time_2013,liu_3d_2016,pan_dense_2019}, whose output representation of the environment remains complicated. On the other hand, SLAM, especially indoor SLAM, utilizes planes as landmarks to achieve the simultaneous motion estimation and environment mapping~\cite{lee_indoor_2012,geneva_lips_2018,yang_monocular_2019,zuo_lic-fusion_2020,sun_plane-edge-slam_2021}. However, 3D reconstruction approaches keep using 3D point clouds to represent environment features and require high computational and storage cost. SLAM approaches focus more on plane fitting and matching rather than accurately represent the complete planar regions with structured boundary characterization. Neither of them can be directly applied to help with the foothold planning problem for legged robots.





Following the core idea of SLAM, various perception schemes for accurate terrain mapping have been proposed for legged robot robot locomotion recently. Height map based strategies are among the most widely adopted schemes~\cite{kolter_control_2008,magana_fast_2019,villarreal_mpc-based_2020,fankhauser_robot-centric_2014,fankhauser_robust_2018}.
Kolter et al. \cite{kolter_control_2008} presented a planning architecture that first generates the height map from a 3D model of the terrain. Such an idea has been extended in~\cite{fankhauser_robot-centric_2014,fankhauser_robust_2018}, where robot-centric elevation maps of the terrain from onboard range sensors are constructed. Excellent experimental results reported in these works have proven that projecting the point clouds onto the discrete height map is an efficient way to represent rough terrain. Nonetheless, all these scheme require an additional step of constructing an associated cost map for foothold planning, which is non-trivial and time consuming. Furthermore, building the height map does not extract the direct information required for legged robot locomotion, and the grid structure of height maps imposes limitation to subsequent foothold choosing. 



The planar region based terrain characterization has also been studied in the literature. Numerous methods for high speed multi-plane segmentation with RGB-D camera readings have been proposed recently \cite{dong_efficient_2018,hulik_fast_2012,proenca_fast_2018,feng_fast_2014,roychoudhury_plane_2021}. Feng et al. \cite{feng_fast_2014} applied Agglomerative Hierarchical Clustering (AHC) method on the graph constructed by groups of point clouds for plane segmentation. Proença and Gao \cite{proenca_fast_2018} presented a cell-wise region growing approach to extract planes and cylinders segments from depth images. These real-time plane segmentation methods well suit robotic application, but the limited FoV of depth camera leads to incomplete detection of physical planes, which prevents us from simply extracting planar regions in single depth measurement.

Building upon the idea of planar region characterization, pioneering efforts on further approximating the planar regions via polygonal or polytopic regions and applying to practical foothold planning schemes have been made. Deits et al.~\cite{akin_computing_2015} presented an iterative regional inflation method to represent safe area by convex polytopes formed by obstacle boundaries. While this approach is efficient, a hand-selected starting point is needed to grow the inscribed ellipsoid of obstacles. Moreover, it is hard to represent complex obstacle-free regions by the convex polytope obtained by inflating inscribed ellipsoid. Bertrand et al.~\cite{bertrand_detecting_2020} constructed an octree to store the point clouds from LIDAR readings, and grouped the nodes into planar regions which are then used for footstep planning of humanoid robots. This work effectively extract horizontal planar regions from LIDAR scans, but the algorithm can only achieve about 2 Hz frame rate and the resulting representation is not guaranteed to be convex, limiting its applications in reactive foothold planning for legged locomotion.

\subsection{Contributions}
The main contributions of this paper are summarized as follows. First, originate from the specific application of foothold planning for legged robots, the proposed approach characterizes terrain environment via polytopic representation that ensures direct applicability to legged locomotion. Second, by combining plane segmentation and merging techniques, the proposed approach reconstructs the complete terrain with limited field of view (FOV), effectively removing various restrictions on the used sensory system and hence expanding the applicability of the proposed approach. Third, through careful integration of the participating modules, the proposed approach is very efficient, capable of running at a higher frequency than the intrinsic frame-rate of the sensory system in our experimental tests, fulfilling the requirement for subsequent planning for legged locomotion problems.

\begin{figure*}[tbp]
\centering
\includegraphics[height=6cm]{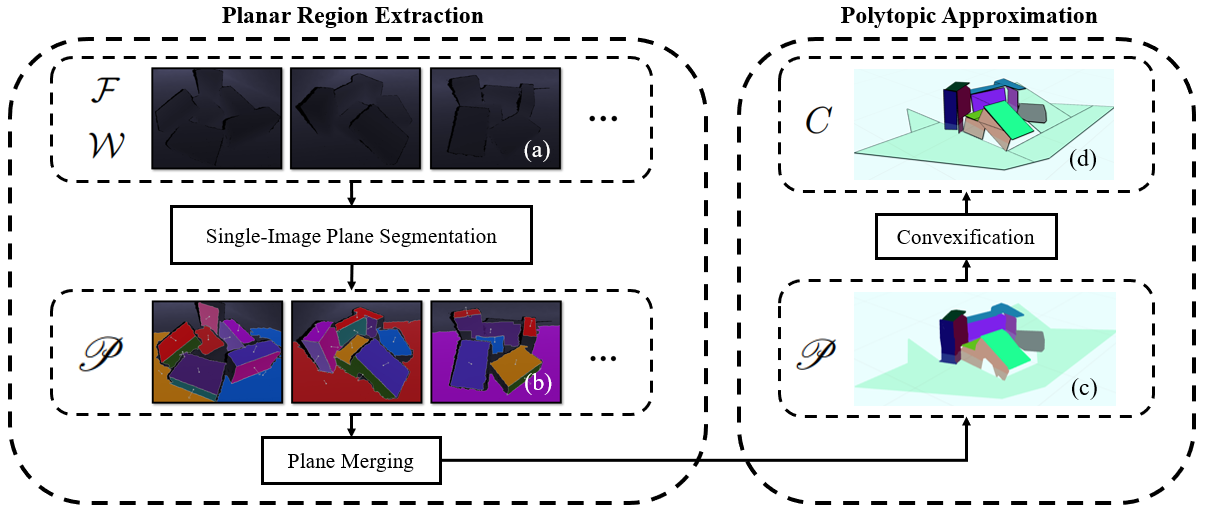}
\caption{Overview of the proposed method. (a) is a sequence of depth images $\F$ with associated camera frames $\W$ (b) illustrates the planes $\mathscr{P}$ segmented in each frame. After plane merging, the extracted planar regions in the world frame is shown in (c). Black lines in (d) show the boundaries of the convex polytopes $C$ obtained by convexification.}
\label{fig:overview}
\end{figure*}

\section{Problem Description and Solution Overview}
\label{Problem Description and Overview of the Approach}

The primary objective of the planar region identification for legged locomotion is to acquire a set of polytopes characterizing feasible supports for foothold selection. In this paper, we particularly focus on the case where only depth/point cloud measurements are available. 

To rigorously formulate the planar region identification problem, we consider a sequence of depth image measurements denoted by $\F = \{F_1,F_2,\ldots, F_N\}$ where $F_i\in \R^{U\times V}, \ \forall i=1,\ldots,N$ denotes the depth image of the $i$-th measurement, and a sequence of associated camera frames $\W = \{W_1,W_2,\ldots,W_N\}$ expressed in an inertial frame. The planar region identification problem studied in this paper aims to construct the collections of low-dimensional polytopic regions $\mathscr{P} = \{\P_1,\P_2,\ldots,\P_N\}$ with $\P_i = \{P_i^1,\ldots,P_i^M\}$ where $P_i^k\subset \R^3, \ \forall k=1,\ldots,M$ is a polytope in the collection. For a generic polytope $P$ considered in this paper, we adopt the following representation
\eq{\label{eq:poly_rep} P = (\V, N, \bar{p}, \vec{n}, \MSE).} In this representation, $\V$ and $\vec{n}$ jointly specifies the polytope, where $\V$ is the set of vertices of the polytope and $\vec{n}$ denotes the normal vector of the two-dimensional polytope in three-dimensional space. The quantities $N, \bar{p}$ and $\MSE$ are introduced to relate the polytope with the point cloud measurement, where $N$ is the number of points associated with the polytope, $\bar{p}$ is the average of all points associated with the polytope, and $\MSE$ is the mean square error between the sampled points and the identified polytope. 

Given the above setup, the planar region extraction and convexification problem studied in this paper can be rigorously formulated below.

\begin{problem}
Given the sequence of depth image measurements $\F$ and the associated camera frames $\W$, find the polytopic representation $\P$ for all planar regions contained in the overall perceived environment. 
\end{problem}

\begin{remark}
Practically speaking, the sequence of depth images is commonly indexed by time as well, making the identification problem an estimation problem in nature. Such a viewpoint is widely adopted in the simultaneous localization and mapping (SLAM) community. In essence, the problem we study in this paper can be viewed as a special SLAM problem with pre-specified structures of the map characterization.  
\end{remark}

To address the planar region identification problem described above, we adopt a two-stage strategy in this paper. At the first stage, the planar regions appeared in a sequence of depth images are identified and merged. Typically, these identified planar regions are expressed with pixel-wise representation, which is overly complicated and not utilizable for applications in foothold planning for legged robots. In view of these issues, we 
subsequently approximate the extracted planar regions via polytopic representations at the second stage, which eventually gives the desired polytopic representations of all candidate planar regions in the perceived environment. 


It is worth noting that, finding all possible planar regions from depth image or point cloud is a combinatorial problem, which is fundamentally challenging. Furthermore, the complexity of real world scenarios makes it practically impossible to perfectly classify all points in the point cloud to some polytopic region. In view of these theoretical and practical difficulties, our proposed solution leverages the special structure of our polytopic representation that is efficient and reliable. In the following subsection, an overview of the proposed solution is first provided. 

\subsection{Overview of the Proposed Framework}
The schematic diagram of the proposed solution is depicted in Fig.~\ref{fig:overview}. As briefly mentioned before, the overall solution consists of two major parts. The planar region detection module takes the raw depth image measurements $\F$ and the associated camera frames $\W$ as inputs. These inputs are first transformed into a point cloud representation with neighborhood information encoded. With this specialized point cloud data, classical multi-plane segmentation techniques can be applied to acquire both the mask/boundary information and the parameters (including center $\bar{p}$, normal vector $\vec{n}$ and mean square error $\MSE$) of the candidate planes from one depth image. As we receive upcoming depth images, the candidate planes segmented from different images are transformed to a common inertial frame through the camera frames $\W$. In order to improve the integrity and accuracy of planar regions representation, newly extracted planes are fused with historical planar regions. For coplanar and connected planes, we first merge their parameters, then rasterize them into a single 2D plane and merge their boundaries by constructing their masks. 
Once the extraction of planar regions are completed, we conservatively approximate the planar regions by a set of convex polygons $C=\{\vec{n},\V_C\}$. Such an approximation not only reduces the complexity of representing a polygonal region, but also offers tractability for future foothold planning schemes. Finally, after converting all convex polygons into the robot's local frame $R$, a robot-centric polytopic planar regions map that can be used in foothold planning is constructed. 


\section{Planar Region Extraction}
\label{Planar Region Extraction}
To extract all planar regions from various depth images of environment, we first need to identify all planes in a single depth image. Then, the identified planar regions belonging to a common physical plane are merged to obtain the a minimum representation. By adopting an efficient cell-based region growing algorithm, all planes in one frame are labelled as point clusters in real time. We then compute the plane parameters and extract the  boundaries of the planar regions. After that, we store the planes with low-dimensional representation in a fixed frame. For all planes in the frame detected from different depth images, we proposed a rasterization-based method to efficiently merge newly extracted plane segments and restore the original planar region in the physical world. In the rest of this section, the details on these two major steps for plane extraction are presented.

\subsection{Single-Image Plane Segmentation}\label{Planar Region Extraction-A}
Single-image plane segmentation is a classical problem that has been extensively studied in the computer vision literature. For applications in robotics, various practical restrictions call for specialized treatment on this problem. Taking into account of the real-time implementable requirement, we follow the idea proposed in~\cite{proenca_fast_2018}. 

Given a frame of depth image captured by RGB-D camera, the method first generates the organized point clouds data segmented into grid cells. Then, Principal Component Analysis (PCA) are applied in each cell to compute the principle axis of the 3D point cluster and accomplish plane fitting. The seed cells are selected if their mean square errors (MSE) are lower than a prescribed threshold and there are no discontinuities inside the cells. Given the seed cells, cell-wise region growing is performed in the order determined by the histogram of planar cell normals. Note that, by utilizing the grid structure of the point clouds in image format, normal vector computing and region growing are operated on point clusters, which significantly reduces the computational cost and therefore accelerates the processing. After cell-wise region growing, the coarse cell-level boundaries of the obtained planes are then refined by performing pixel-wise region growing along the boundary extracted by morphological operations. With the point-wise refinement, the accuracy of segmented plane boundary is greatly improved.

Now, every plane in the camera frame is labeled as a point cluster $\{p_i\}_{i=1}^k \subset{\mathbb{R}^3}$. The normal vector $\vec{n}$ is the cross product of the first two principal axis computed through PCA. The centroid of the plane is defined as: 
 $\bar{p} = \frac{1}{k} \cdot \sum_{i=1}^k{p_i}$
The mean square error is calculated by:
$$ \MSE = \frac{1}{k} \sum_{i=1}^k{(\vec{n} \cdot p_i - b)^2} $$
where $b = \vec{n} \cdot \bar{p}$ is a bias element of the plane.

With labeled pixels in the camera space, a digitized binary 
image is obtained for each plane. We then extract the boundary of each plane in the camera space. Given the vertices $\V_{img}=\{u_i,v_i\}_{i=1}^p$ of the 2D contour extracted in the image, 3D vertices $\V_{cam}=\{x_i,y_i,z_i\}_{i=1}^p$ in the camera frame can be calculated through the prospective camera model and the plane equation:
$$ \begin{cases}
x = z(u-c_x)/f \\ y = z(v-c_y)/f \\ z = \frac{\vec{n} \cdot \bar{p}}{n_x(u-c_x)/f+n_y(v-c_y)/f+n_z}
\end{cases} $$
where $K=\begin{bmatrix} f&0&c_x\\0&f&c_y\\0&0&1 \end{bmatrix}$ is the intrinsic matrix of the camera, $\vec{n}=(n_x,n_y,n_z)$ is the normal vector of the plane.

For all planes extracted in one frame, the vertices $\V_{cam}$, normal vector $\vec{n}$, and centroid $\bar{p}$ in the camera space can be directly transferred to the world frame given the corresponding camera pose.


\begin{figure}[t]
\centering
\includegraphics[height=4cm]{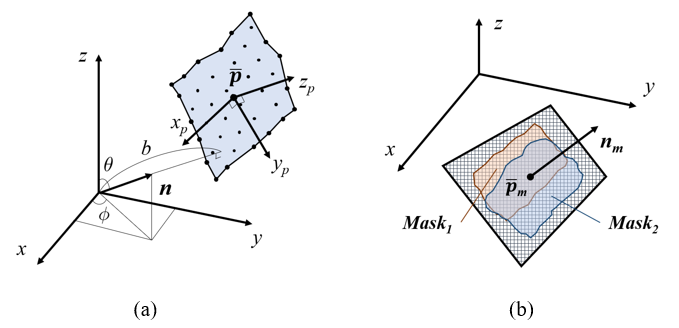}
\caption{(a) Coordinate frame and plane parameters. (b) Definition of plane masks.}
\label{fig:rasterization}
\end{figure}

\subsection{Plane Merging}
Due to the limited field of view (FOV) of the depth camera, planes detected in each frame during robot motion could be only a part of the original plane. Once the planes segmented from different depth images with different camera frames are obtained, we need to merge those that actually correspond to the same physical plane to restore the real planar region and to ensure the integrity and accuracy of planar region detection. 


\begin{algorithm}[t]
    \caption{Plane Merging}
	\label{alg:Plane Merging}
	\KwIn{\par List of historical planes $L_H$ and new planes $L_N$}
    \KwOut{$L_H$ after merging}
	\BlankLine
	\ForEach{$\P_N \in L_N$}{
	    \ForEach{$\P_H \in L_H$}{
	        \If{isCoplanar($\P_N$, $\P_H$)}{
	            $Para_M$ $\leftarrow$ MergeParameter($\P_N$, $\P_H$)\par
	            $Mask_N$ $\leftarrow$ Rasterize($\P_N$, $Para_M$)\par
	            $Mask_H$ $\leftarrow$ Rasterize($\P_H$, $Para_M$)\par
	            \If{$Mask_N \cap Mask_H \not\in \emptyset$}{
	                $Mask_M$ = $Mask_N \cup Mask_H$\par
	                $\P_M$ = InvRasterize($Mask_M$, $Para_M$)\par
	                $L_H$.Replace($\P_H$, $\P_M$)\par
	                $L_N$.Delete($\P_N$)\par
	                $\P_N$ = $\P_M$
	            }
	        }
	    }
	}
	$L_H$.Append($L_N$)
\end{algorithm}

After plane extraction, all planes extracted in a frame are stored in the world frame as a global map $\mathscr{P}$ with their vertices ($\V_P=\{x_i,y_i,z_i\}_{i=1}^p$), point number ($N$), mean ($\bar{p}$), normal vector ($\vec{n}$) and mean square error ($\MSE$): 
$$\mathscr{P} = \{\P_i = (\V_P,N,\bar{p},\vec{n},\MSE)\}_{i=1}^n$$
For a newly extracted plane, we first find the planes coplanar with it by following criterion:
$$ \begin{cases}
\mid n_1 \cdot n_2 \mid < \tau_{\theta} \\
\mid n_1 \cdot \bar{p}_1 - n_2 \cdot \bar{p}_2\mid < \tau_{b}
\end{cases} $$
where $\tau_{\theta}$ and $\tau_{b}$ are two preset thresholds.\par
Centroids of planes after merging are updated through the following formulas:
\eqn{\ald{{N}_m& = \sum_i{{N}_i}, \quad \bar{p}_m&=\frac{1}{{N}_m} \cdot \sum_i{\bar{p}_i \cdot {N}_i}} }
The low dimensional plane representation can significantly reduce the computational and storage cost. However, as we abandon the original point clouds data for each plane, the normal vectors of the merged planes can not be directly obtained. To account for this issue, we notice that the MSE reflects the accuracy of plane fitting model. Therefore, we take it as a weight to fuse normal vector $\vec{n}$ of coplanar planes. By transferring $\vec{n}$ to the spherical coordinate system ($\vec{n}=[\theta,\phi,1]$), we merge $[\theta,\phi]$ according to the MSE as follows:
\begin{gather*}
\MSE_m = (\sum_i{\MSE_i^{-1}})^{-1} \\
\begin{bmatrix} \theta_m\\\phi_m \end{bmatrix} = \MSE_m \cdot \sum_i{\MSE_i^{-1} \cdot \begin{bmatrix} \theta_i\\\phi_i \end{bmatrix}}
\end{gather*} \par
Once the parameters are updated, we first transfer the contour $\V_P$ to the coordinate system with $\bar{p}_m$ as origin and $n_m$ as the z-axis (See Fig \ref{fig:rasterization}). By rasterizing the x-y plane of the coordinate and fill the grids inside the boundaries, we then obtained the mask of each plane in 2D matrix form. Given the masks of all planes, we can simply check their connectivity through bit-wise-and and merge connected planes with bit-wise-or operation. With the merged mask in binary matrix form, its vertices can be extracted again as in Section \ref{Planar Region Extraction-A}. When the 2D vertices are transferred back to the world frame by inverse transformation, we obtain the digitized boundary $\V_P$ of the merged plane. The overall merging algorithm is outlined in Algorithm \ref{alg:Plane Merging}.

\section{Polytopic Approximation}
\label{Polytopic Approximation}

Through combining a plane segmentation step and a merging step, we have successfully segmented candidate planes given a sequence of depth images and their corresponding camera frames. However, the extracted planar regions are expressed via pixel-wise mask matrices, which does not imply direct applications to foothold planning problems for legged locomotion. Concerned with the potential applications in legged locomotion, we need to further simplify the representation of the planar regions via polytopes. In the following section, we develop our approach to approximate the constructed planar regions via polytopes efficiently and accurately.

\subsection{Contour Simplification}
After plane extraction, the accurate high-resolution contour of the planar region are obtained. However, the large number of vertices in plane representation may contain much detailed information which is nonessential for foothold planning. Besides, numerous small notches in the contour may result in a large number of convex polygons and significantly increase the computational and storage cost. Thus, the high-resolution contour needs to be simplified while keeping the original shape of the planar area and ensuring the safety of footstep planning.

\begin{figure}[t]
\centering
\includegraphics[height=4cm]{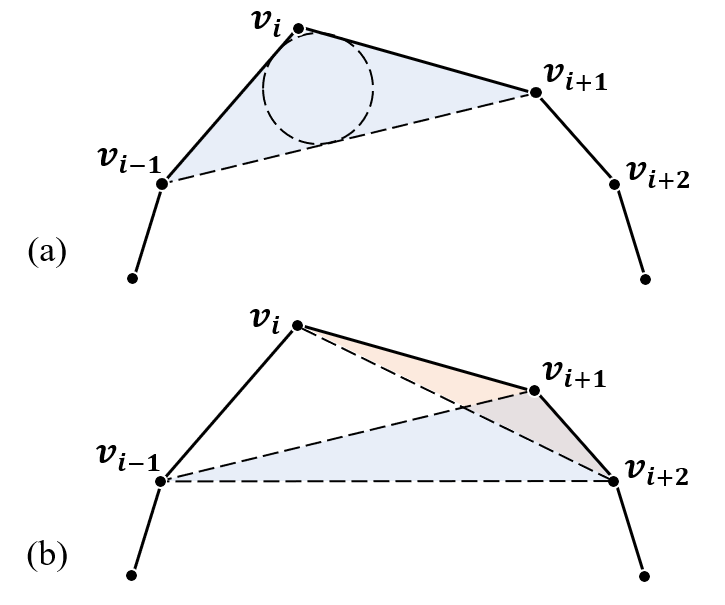}
\caption{(a) Triangle defining the error $\varepsilon$ associated with the elimination of point $v_i$, and its inscribed circle with diameter $d$. (b) Triangles defining $\varepsilon$ of point $v_{i+1}$ before (red) and after (blue) the elimination of $v_i$.}
\label{fig:approximation sketch}
\end{figure}

\begin{algorithm}[b]
    \caption{Contour simplification}
	\label{alg:Contour Simplification}
	\KwIn{ Plane Vertices $\V_P = \{v_i=(x,y,z)\}_{i=1}^p$}
	\BlankLine
	Compute $\varepsilon$ for all $v_i \in \V_P$\par
	$H \leftarrow$ BuildMinHeap($\V_P$, $\varepsilon$)\par
	\Repeat{$\varepsilon_i > \varepsilon$}{
        $v_i,\varepsilon_i \leftarrow$ $H$.ReturnMinVertex()\par
        \If{isConcave($v$)}{
            $d \leftarrow$ ComputeInnerCircleDiameter($v_i$)\par
            \If{$d > d_{r}$}{
            $H$.Update($v_i$, Infinity)\par
            continue
            }
        }
        $H$.UpdateNeighborVertex($v_i$)\par
        $H$.Delete($v_i$)
    }
\end{algorithm}
To simplify the contour, we first build a min-heap data structure from the contour $\V_P$. The value of each leaf $v_i \in \V_P$ is defined as the area of the triangle formed by vertices $v_{i-1}$, $v_i$ and $v_{i+1}$. Then, we iteratively pop out the vertex with minimum value until a preset threshold $\varepsilon$ is reached. After popping out vertex $v_i$, the value of $v_{i-1}$ and $v_{i+1}$ should be updated by their new neighbor vertices. In this way, the vertex whose elimination introduces least error $\varepsilon$ is greedily deleted, and we can control the roughness of the contour by adjusting the threshold $\varepsilon$.\par
However, the planar region could be dilated through this method if concave vertices are deleted. The larger boundary may cause the footstep being selected outside the real plane. To avoid this situation, we check the convexity of each vertex before vertex popping. For concave vertex, we calculate the diameter $d$ of the inscribed circle of the triangle formed by vertices $v_{i-1}$, $v_i$ and $v_{i+1}$. If the value is smaller than the diameter of the robot's foot $d_{r}$, vertex $v_i$ is deleted. Otherwise, $v_i$ is preserved and moved to the bottom of the heap. In this way, the planar region can be approximated safely without causing the robot to miss its step. The adapted method is described in Algorithm \ref{alg:Contour Simplification}. For time complexity, the use of the min-heap data structure yields a worst-case complexity of $O(n\log n)$.

\subsection{Convex Partition}
Once Contour Simplification is obtained, we divide the planar region into multiple convex polygons $C=\{\vec{n},\V_C\}$ through a vector method. Given the contour $\V_P$ of a plane, we iterate through each vertex $v_i$ and check its convexity by the sign of the cross product $(v_i-v_{i-1}) \times (v_{i+1}-v_i)$. For a concave vertex $v_i$, we extend the vector $v_i-v_{i-1}$ and compute the first intersection $s$ with the boundary. The polygon is then divided in two parts by the segment $v_is$. The procedures are repeated on the resulting two contours until no concave vertex is found.\par
The method is simple but efficient, a polygon with $p$ vertices and $q$ concave vertices can be divided into no more than $q+1$ convex polygons with time complexity $O(p+q)$. In practice, the computation time of convex partition can be nearly neglected since concave vertices become much fewer after Contour Simplification.

\begin{figure}[t]
\centering
\includegraphics[height=3cm]{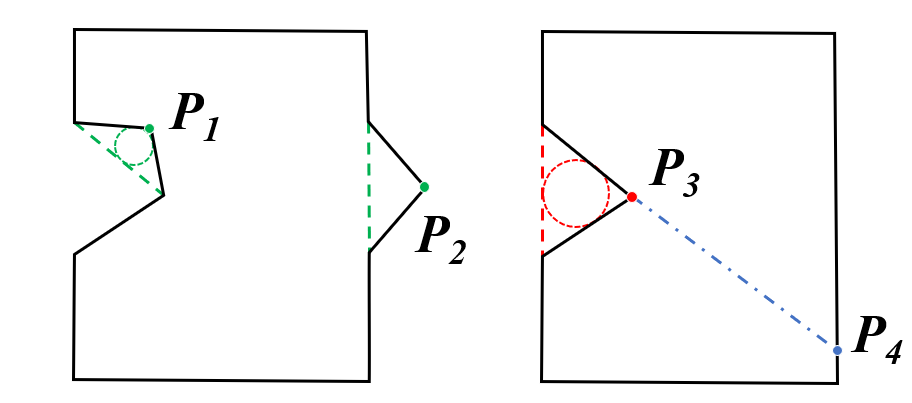}
\caption{Plane contour can be simplified and decomposed into convex polygons by proposed method. $P_1$: concave vertex to be deleted satisfying $\varepsilon_i < \varepsilon$ and $d < d_{r}$. $P_2$: convex vertexto be deleted satisfying $\varepsilon_i < \varepsilon$. $P_3$: preserved concave vertex with $\varepsilon_i < \varepsilon$ but $d > d_{r}$. $P_4$: intersection between contour and extended concave vector. The plane is decomposed to two convex polygons by segment $P_3P_4$.}
\label{fig:convex approximation sketch}
\end{figure}

\begin{figure}[t]
\centering
\includegraphics[width=6cm]{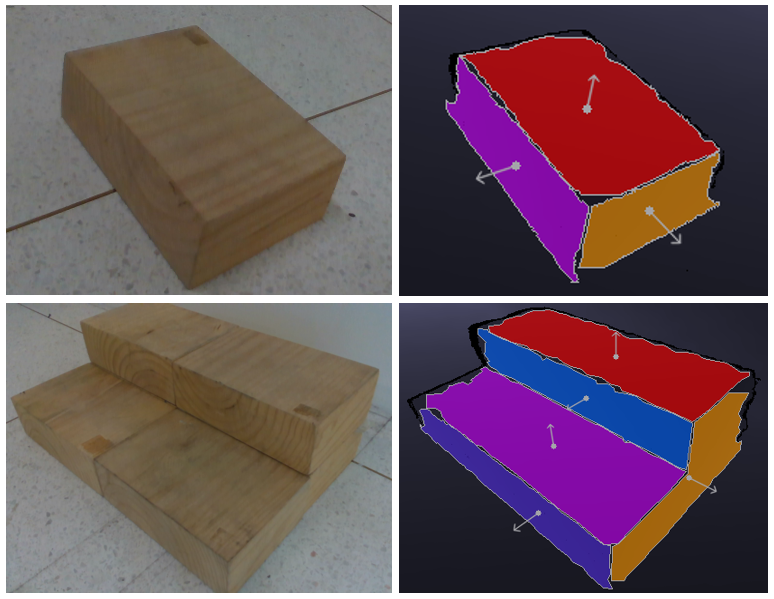}
\caption{Results of single-image plane extraction. Normal vectors and centroids of the planes are indicated by white arrows and points.}
\label{fig:single frame}
\end{figure}
\begin{figure}[t]
\centering
\includegraphics[width=6cm]{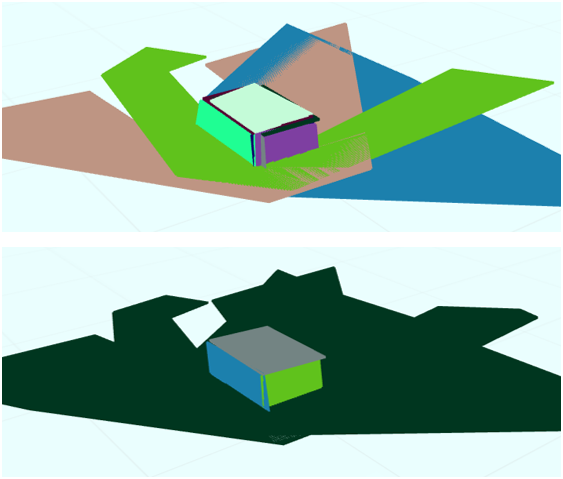}
\caption{Planes extracted from three images captured at different viewpoints before and after merging.}
\label{fig: plane merging}
\end{figure}

\begin{figure}[t]
\centering
\includegraphics[width=5cm]{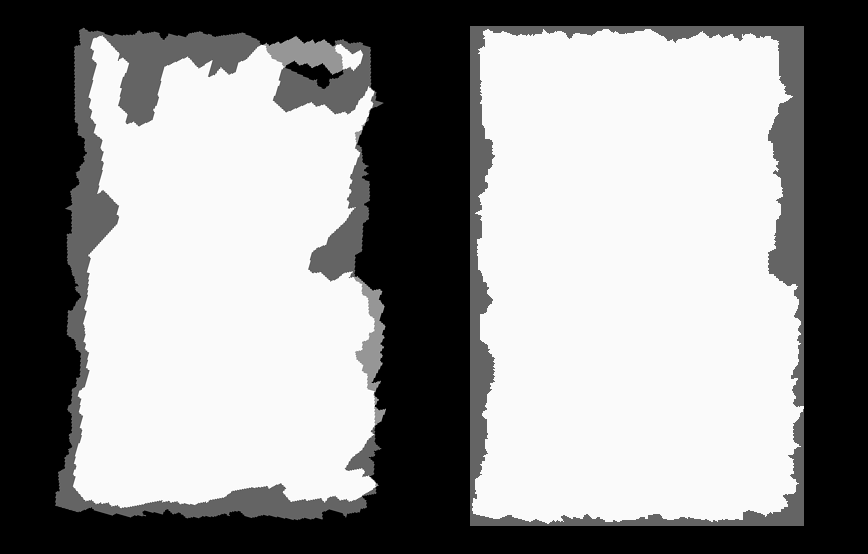}
\caption{Left: two planes before merging (IoU = 86.3$\%$ and 73.0$\%$). Right: ground truth and the plane after merging (IoU = 87.4$\%$).}
\label{fig:plane merging example}
\end{figure}

\begin{figure*}[t]
\centering
\includegraphics[height=4cm]{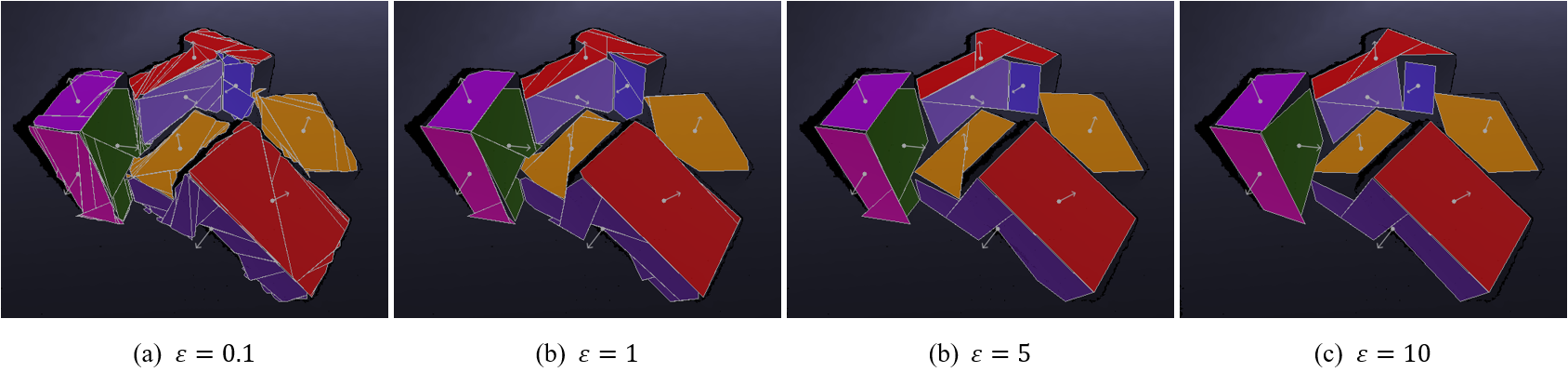}
\caption{Results of polytopic approximation with different $\varepsilon$ (cm$^2$).}
\label{fig:convex approximation}
\end{figure*}

\begin{table*}[tp!]
\centering
\begin{tabular}{|l|ccc|ccc|ccc|c|}
\hline
Scene                  &                           & 1                         &      &                           & 2                         &      &                           & 3                         &      & Average       \\ \hline
Frame                  & \multicolumn{1}{c|}{1}    & \multicolumn{1}{c|}{2}    & 3    & \multicolumn{1}{c|}{4}    & \multicolumn{1}{c|}{5}    & 6    & \multicolumn{1}{c|}{7}    & \multicolumn{1}{c|}{8}    & 9    & \textbf{}     \\ \hline
$\alpha$($^{\circ}$)   & \multicolumn{1}{c|}{1.52} & \multicolumn{1}{c|}{1.27} & 2.32 & \multicolumn{1}{c|}{1.43} & \multicolumn{1}{c|}{1.13} & 0.81 & \multicolumn{1}{c|}{1.75} & \multicolumn{1}{c|}{1.19} & 0.59 & \textbf{1.33} \\ \hline
$\Delta b$ (mm)        & \multicolumn{1}{c|}{16.5} & \multicolumn{1}{c|}{19.9} & 20.6 & \multicolumn{1}{c|}{30.3} & \multicolumn{1}{c|}{18.4} & 14.6 & \multicolumn{1}{c|}{28.7} & \multicolumn{1}{c|}{17.3} & 15.2 & \textbf{20.2} \\ \hline
IoU (\%)               & \multicolumn{1}{c|}{85.4} & \multicolumn{1}{c|}{71.9} & 81.5 & \multicolumn{1}{c|}{76.8} & \multicolumn{1}{c|}{82.3} & 81.8 & \multicolumn{1}{c|}{71.5} & \multicolumn{1}{c|}{77.5} & 85.3 & \textbf{79.3} \\ \hline
$\alpha_m$($^{\circ}$) &                           & 1.33                      &      &                           & 1.14                      &      &                           & 1.15                      &      & \textbf{1.21} \\ \hline
$\Delta b_m$ (mm)      &                           & 18.1                      &      &                           & 20.1                      &      &                           & 16.4                      &      & \textbf{18.2} \\ \hline
IoU$_m$ (\%)           &                           & 87.7                      &      &                           & 86.3                      &      &                           & 90.2                      &      & \textbf{88.1} \\ \hline
\end{tabular}
\caption{Results of plane extraction and plane merging test.}
\label{tab:experiment data}
\end{table*}

\begin{figure}[t]
\centering
\includegraphics[width=7cm]{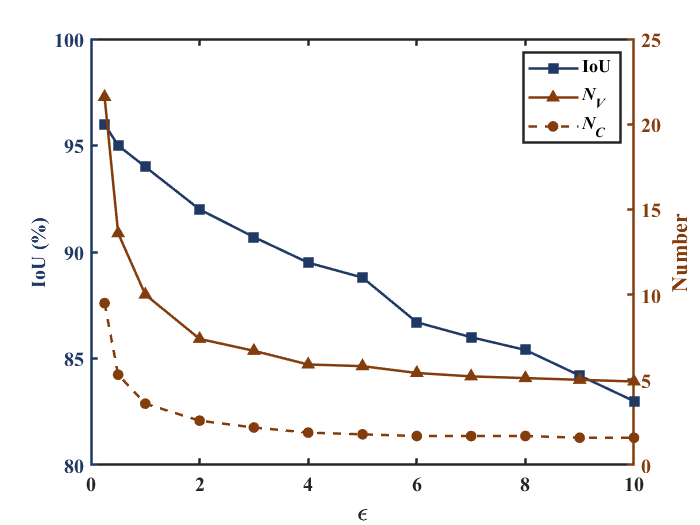}
\caption{Results of polytopic approximation with different value of $\varepsilon$. $N_V$: average vertices number per plane. $N_C$: average convex polygons number per plane. IoU: Intersection over Union of the planes before and after polytopic approximation.}
\label{fig:convex approximation value}
\end{figure}
\begin{figure}[tbp]
\centering
\includegraphics[width=7cm]{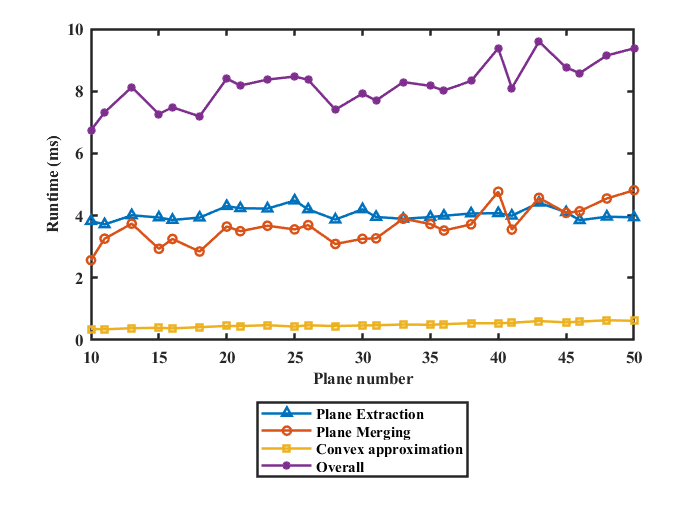}
\caption{Run-time of the algorithm.}
\label{fig:runtime}
\end{figure}

\begin{figure}[tbp]
\centering
\includegraphics[width=6cm]{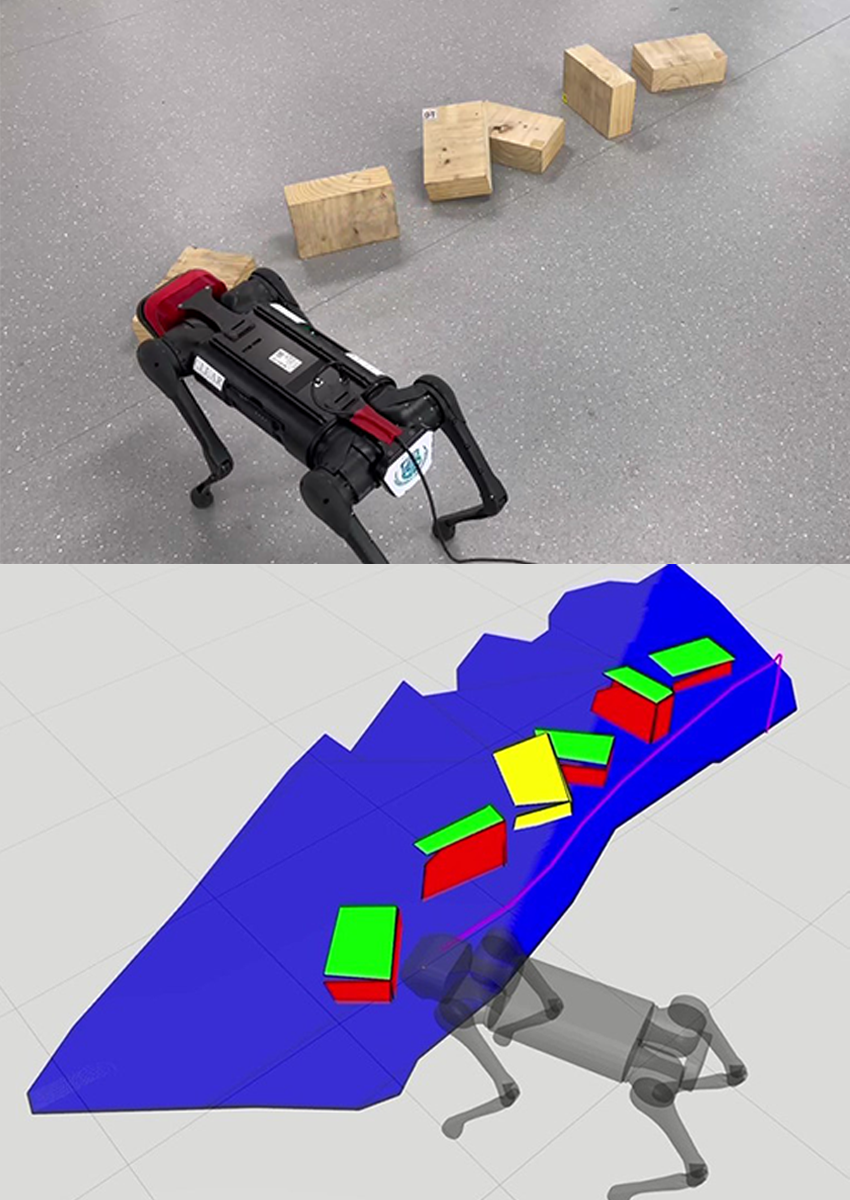}
\caption{Moving camera result on a quadrupedal robot. The upper figure is real scenario. All the detected planes are illustrated in lower figure. Blue plane is ground. Green planes are horizontal. Red planes are vertical. Yellow stands for inclined planes. Violet line is robot trajectory.}
\label{fig:moving camera}
\end{figure}

\section{Experimental Results}
\label{Experimental Results}

To comprehensively evaluate the proposed method in terms of accuracy, robustness and efficiency, we conducted three sets of experiments as described in following subsections. The depth images are captured by an Intel Realsense D435 RGB-D camera under VGA resolution ($640 \times 480$ pixels). All algorithms are implemented in C++ and run on a laptop with Intel i5-7300HQ CPU.

\subsection{Planar Region Extraction Result}
We first evaluated the plane extraction algorithm on a single frame as illustrated in Fig. \ref{fig:single frame}. The experiment was conducted over 3 scenarios with multiple planar regions. For each case, we captured 3 frames at different viewpoints with known camera pose. The proposed plane extraction algorithm segmented disconnected planes in each frame and output their normal vectors $\vec{n}$, centroids $\bar{p}$, and vertices $\V_p$. Given the ground truth parameters, we evaluated the accuracy of the algorithm by the average angle $\alpha$ between the normal vectors, average difference of the bias element $b = \vec{n} \cdot{\bar{p}}$, and the Intersection over Union (IoU) after the measured plane was projected to the corresponding ground truth plane. The results are shown in Table \ref{tab:experiment data}. Considering the original bias of the depth camera (depth accuracy: $<$2\% at 2m), the results are acceptable.

Then, for three frames captured at the same scenario from different view points, we perform plane merging on the planes extracted by single frame segmentation. An intuitive comparison of the planes before and after merging is shown in Fig. \ref{fig: plane merging}. Again, we computed the plane parameters after plane merging and compare with the ground truth as shown in Table~\ref{tab:experiment data}. Compared to the results before plane merging, the average angle bias of the normal vector $\alpha$ and the average difference of plane bias $\Delta b$ are reduced by 9\%. The average IoU is improved by 11\% after plane merging. As shown in Fig. \ref{fig:plane merging example}, for planes that cannot be completely detected in a single frame due to camera pose or sensor bias, plane merging helps to reconstruct the original plane by fusing multiple plane segments extracted from different frames. In practice, the accuracy of the planes after merging is adequate for foothold planning of legged robot.

\subsection{Polytopic Approximation Result}
We tested the proposed polytopic approximation method with different preset threshold $\varepsilon$ as discussed in Section \ref{Polytopic Approximation} B. The planar regions after approximation and all corresponding convex components are illustrated in Fig. \ref{fig:convex approximation}. The average number of vertices and convex polygons number per plane when choosing different $\varepsilon$ are shown in Fig. \ref{fig:convex approximation value}. We also computed the average IoU of the planes before and after polytopic approximation with different threshold. 

The run-time of the overall algorithm and each module is illustrated in Fig. \ref{fig:runtime}. According to the results, as the number of detected planes increases from 10 to 50, the overall running time per frame only increased by around 2ms (7ms to 9ms). In other words, the method can achieve more than 100 Hz frame rate in most cases.

\subsection{Moving Camera Result}
To evaluate the overall algorithm, we mounted the depth camera on a quadruped robot with a visual odometry (Realsense T265) to provide pose estimation. The algorithm ran continuously during robot locomotion, and built a global map of the planar region in the surrounding environment as shown in Fig.\ref{fig:moving camera}.


\section{Conclusion}
\label{Conclusion}
In this paper, we present a plane segmentation method for extracting plane structures from depth image to help for legged robots and its foothold planning. To achieve this, we combine different methods to do planes extraction and post-processing. First, all planes are extracted in one frame and stored into low-dimensional representation in global map. Then, limited by FoV, a physical plane may not be fully constructed in a single frame. Also, error may exist between two detected planes for the same physical plane in different frames because of camera movement. Thus, we use a novel plane merging method to fuse two plane features into one. After planes are extracted, a post-processing is implemented. It simplifies the contours of each plane so that the extracted plane features are transformed into polygons. After that, these plane regions are partitioned into convex polygons so that foothold planning method can perform on it directly. All the work in this paper is aim at legged robot and foothold planning, so the output are convex polygons and the efficiency for the whole algorithm is guaranteed. Finally, we illustrate the performance of this method and each part of it, by implementing it in different scenarios.

\bibliographystyle{ieeetr}
\bibliography{mybibfile.bib}

\end{document}